%% file: main.tex
\documentclass[runningheads,a4paper]{llncs}
\usepackage[top=2.5cm, bottom=2.5cm, left=2.5cm, right=2.5cm]{geometry}
\usepackage{amsmath}
\usepackage{amssymb}
\usepackage{graphicx}
\usepackage{booktabs}
\usepackage{listings}
\usepackage{xcolor}
\usepackage{hyperref}
\usepackage{caption}
\usepackage{subcaption}

\usepackage{url}
\usepackage{color}
\usepackage{cite}
\usepackage{epsfig}
\usepackage{float}
\usepackage{tcolorbox}  
\usepackage{tabularx}
\usepackage{arydshln}
\usepackage{ulem}

\usepackage[nomargin,inline,marginclue,draft]{fixme}
\usepackage{cleveref}
\fxsetup{status=draft}
\fxsetup{theme=color, mode=multiuser} \FXRegisterAuthor{me}{ame}{\color{red}Me}
\newcommand{\orcid}[1]{\href{https://orcid.org/#1}{\includegraphics[scale=0.02]{figures/ORCIDiD_icon128x128.jpg}}} 

\hypersetup{
    colorlinks=true
}
\title{Cancer Type, Stage and Prognosis Assessment from Pathology Reports using LLMs}
\titlerunning{Cancer Type, Stage and Prognosis Assessment from Pathology Reports using LLMs}

\input{0_author_list}

\begin{document}
    \mainmatter
    \maketitle
    \setcounter{footnote}{0} 
    \begin{abstract}
    Large Language Models (LLMs) have shown significant promise across various natural language processing tasks. However, their application in the field of pathology, particularly for extracting meaningful insights from unstructured medical texts such as pathology reports, remains underexplored and not well quantified. In this project, we leverage state-of-the-art language models, including the GPT family, Mistral models, and the open-source Llama models, to evaluate their performance in comprehensively analyzing pathology reports. Specifically, we assess their performance in cancer type identification, AJCC stage determination, and prognosis assessment, encompassing both information extraction and higher-order reasoning tasks. Based on a detailed analysis of their performance metrics in a zero-shot setting, we developed two instruction-tuned models: \textbf{Path-llama3.1-8B} and \textbf{Path-GPT-4o-mini-FT}. These models demonstrated superior performance in zero-shot cancer type identification, staging, and prognosis assessment compared to the other models evaluated.
    \end{abstract}   
    
    \input{1_introduction}

    \input{2_results}

    \input{3_discussion}

    \input{4_methods}    

    \section*{Acknowledgments} 
    J.T.R. was supported by a Medical Scientist Training Program grant from the National Institute of General Medical Sciences of the NIH under award no. T32GM152349 to the Weill Cornell/Rockefeller/Sloan Kettering Tri-Institutional MD-PhD Program.
    
    \section*{Data Availability} 
    \label{Data Availability}
    The complete data set and code used to create it are available at - 
    \newline
    \href{https://huggingface.co/datasets/rosenthal/tcga-path-notes}{https://huggingface.co/datasets/rosenthal/tcga-path-notes}.

    \section*{Code Availability} 
    The complete code and the models are available at - 
    \newline
        \href{https://github.com/rachitsaluja/PathRep-Bench}{https://github.com/rachitsaluja/PathRep-Bench}.
    
    \bibliographystyle{ieeetr}
    \bibliography{bibliography.bib}
    \newpage
    \appendix

    \section{Appendix 1: Prompts}
    \label{Appendix-Prompts}

\definecolor{RoyalGray}{RGB}{66, 66, 54}

\begin{figure}[H]
\centering
\begin{tcolorbox}[
    colback=black!5!white,    
    colframe=RoyalGray,      
    title=Prompt for Cancer Type Identification,
    fonttitle=\bfseries,  
    coltitle=white,          
    colbacktitle=RoyalGray,  
    boxrule=1mm,             
    width=\textwidth         
]

\textbf{System Prompt:} 
You are highly knowledgeable and intelligent pathology AI assistant. Your main responsibility is to extract the patient's diagnosis from the text provided and give an answer only from the set of options. You will only output it as a JSON Object and nothing else. Here are the set of options:

'Adrenocortical carcinoma', 'Bladder Urothelial Carcinoma', 'Brain Lower Grade Glioma', 'Breast invasive carcinoma', 'Cervical squamous cell carcinoma and endocervical adenocarcinoma', 'Cholangiocarcinoma', 'Colon adenocarcinoma', 'Esophageal carcinoma', 'Glioblastoma multiforme', 'Head and Neck squamous cell carcinoma', 'Kidney Chromophobe', 'Kidney renal clear cell carcinoma', 'Kidney renal papillary cell carcinoma', 'Liver hepatocellular carcinoma', 'Lung adenocarcinoma', 'Lung squamous cell carcinoma', 'Lymphoid Neoplasm Diffuse Large B-cell Lymphoma', 'Mesothelioma', 'Ovarian serous cystadenocarcinoma', 'Pancreatic adenocarcinoma', 'Pheochromocytoma and Paraganglioma', 'Prostate adenocarcinoma', 'Rectum adenocarcinoma', 'Sarcoma','Skin Cutaneous Melanoma', 'Stomach adenocarcinoma', 'Testicular Germ Cell Tumors','Thymoma', 'Thyroid carcinoma', 'Uterine Carcinosarcoma', 'Uterine Corpus Endometrial Carcinoma', 'Uveal Melanoma'. You have to pick one these options without fail and you cannot print any other text. You as an assistant cannot lie.
\vspace{0.3cm}

\textbf{User Prompt:} What is the diagnosis from this text? Please output it as a JSON object, just generate the JSON object without explanations. ("INSERT PATHOLOGY REPORT")

\end{tcolorbox}
\caption{Prompt for Cancer Type Identification}
\label{Fig7-Prompt-Cancer Type}
\end{figure}

\vspace{-0.5cm}

\definecolor{RoyalGray}{RGB}{66, 66, 54}

\begin{figure}[H]
\centering
\begin{tcolorbox}[
    colback=black!5!white,    
    colframe=RoyalGray,      
    title=Prompt for AJCC Stage Identification,
    fonttitle=\bfseries,  
    coltitle=white,          
    colbacktitle=RoyalGray,  
    boxrule=1mm,             
    width=\textwidth         
]

\textbf{System Prompt:} 
You are an expert medical pathology AI assistant. You are provided with a question about which stage of cancer does the patient have along with the patient's pathology report with multiple answer choices.

Your goal is to think through the question carefully and explain your reasoning step by step before selecting the final answer as a JSON.
Respond only with the reasoning steps and answer as specified below.
Below is the format for each question and answer:

Input:
\#\# Question: \{\{question\}\}

\{\{answer choices\}\}

Output:
\#\# Answer

(model generated chain of thought explanation)

Therefore, the answer is - 

\{"stage" : ANSWER (e.g. Stage I, Stage II, Stage III, Stage IV)\}

\vspace{0.3cm}

\textbf{User Prompt:} Can you identify the AJCC Stage of the Cancer from the following Pathology Report? ("INSERT PATHOLOGY REPORT") 

Options - 

(A) Stage I 

(B) Stage II 

(C) Stage III 

(D) Stage IV

\end{tcolorbox}
\caption{Prompt for AJCC Stage Identification}
\label{Fig8-Prompt-AJCC Stage}
\end{figure}

\newpage

\definecolor{RoyalGray}{RGB}{66, 66, 54}

\begin{figure}[H]
\centering
\begin{tcolorbox}[
    colback=black!5!white,    
    colframe=RoyalGray,      
    title=Prompt for Prompt for Prognosis Assessment,
    fonttitle=\bfseries,  
    coltitle=white,          
    colbacktitle=RoyalGray,  
    boxrule=1mm,             
    width=\textwidth         
]

\textbf{System Prompt:} 
You are an expert medical pathology AI assistant. You are provided with a question whether a patient will surivive after a particular given time or not along with the patient's pathology report with multiple answer choices.

Your goal is to think through the question carefully and explain your reasoning step by step before selecting the final answer as a JSON.

Respond only with the reasoning steps and answer as specified below.

When answering user questions follow these examples-

Can you determine if the patient will survive after ("INSERT MEAN TIMES") years from the following Pathology Report? ("INSERT SUMMARIZED PATHOLOGY REPORT")

Options - 

(A) True 

(B) False

Answer:

\{"Survival": "True"\}

\textbf{PROVIDED WITH 8 EXAMPLES, 4 with True and 4 with False.}

\vspace{0.3cm}

\textbf{User Prompt:} Can you determine if the patient will survive after ("INSERT MEAN TIMES") years from the following Pathology Report? ("INSERT PATHOLOGY REPORT") 

Options - 

(A) True 

(B) False 

\end{tcolorbox}
\caption{Prompt for Prognosis Assessment}
\label{Fig9-Prompt-Prognosis}
\end{figure}

\vspace{-0.5cm}











\section{Appendix 2: Confusion Matrices}

\begin{figure}[H]
      \centering
      \includegraphics[width=1\linewidth]{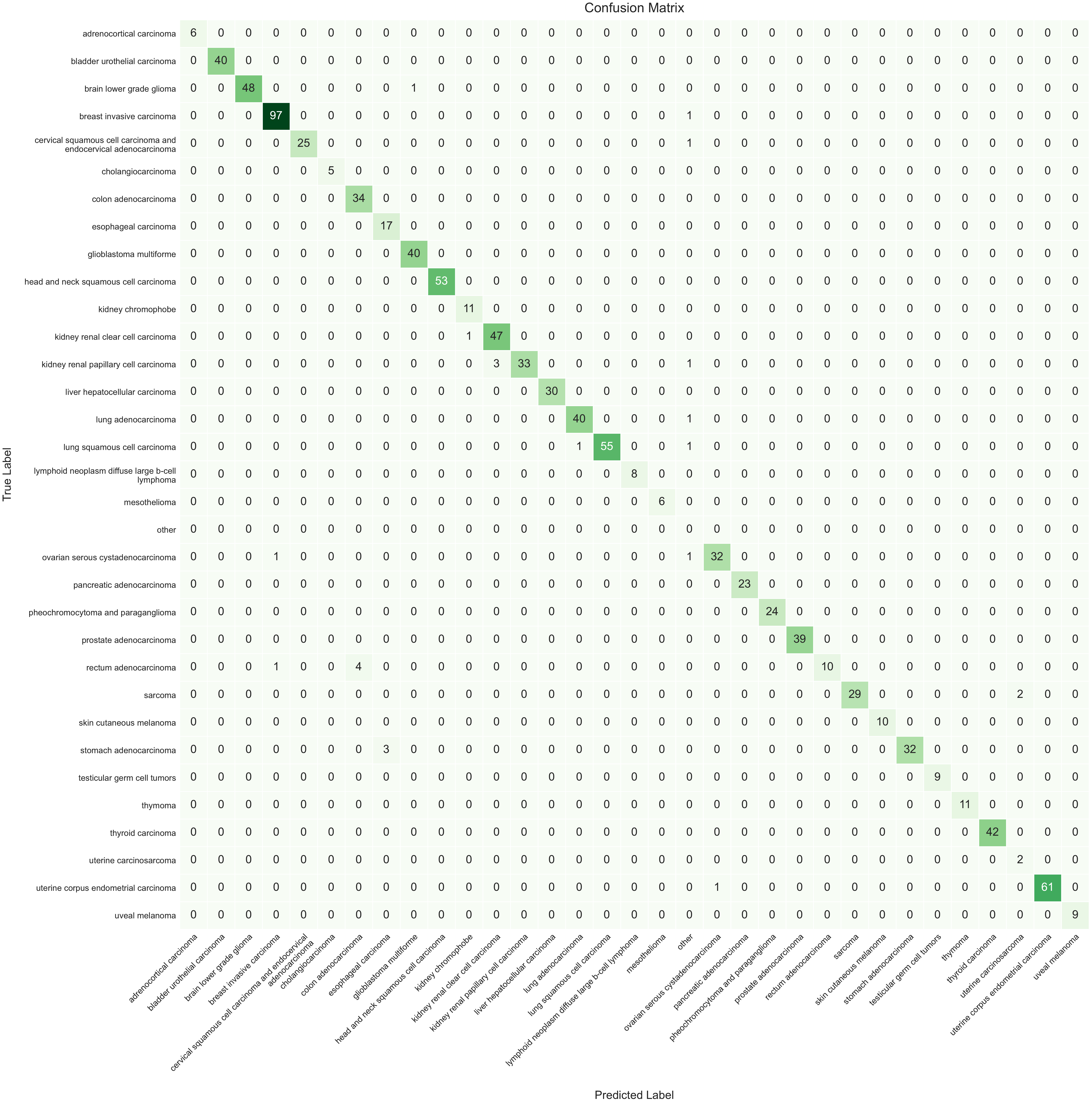}  
      \caption{Cancer Type Identification Confusion Matrix for GPT-4o.}
    \label{Task1-CM-GPT-4o}
\end{figure}

\newpage

\begin{figure}[H]
      \centering
      \includegraphics[width=1\linewidth]{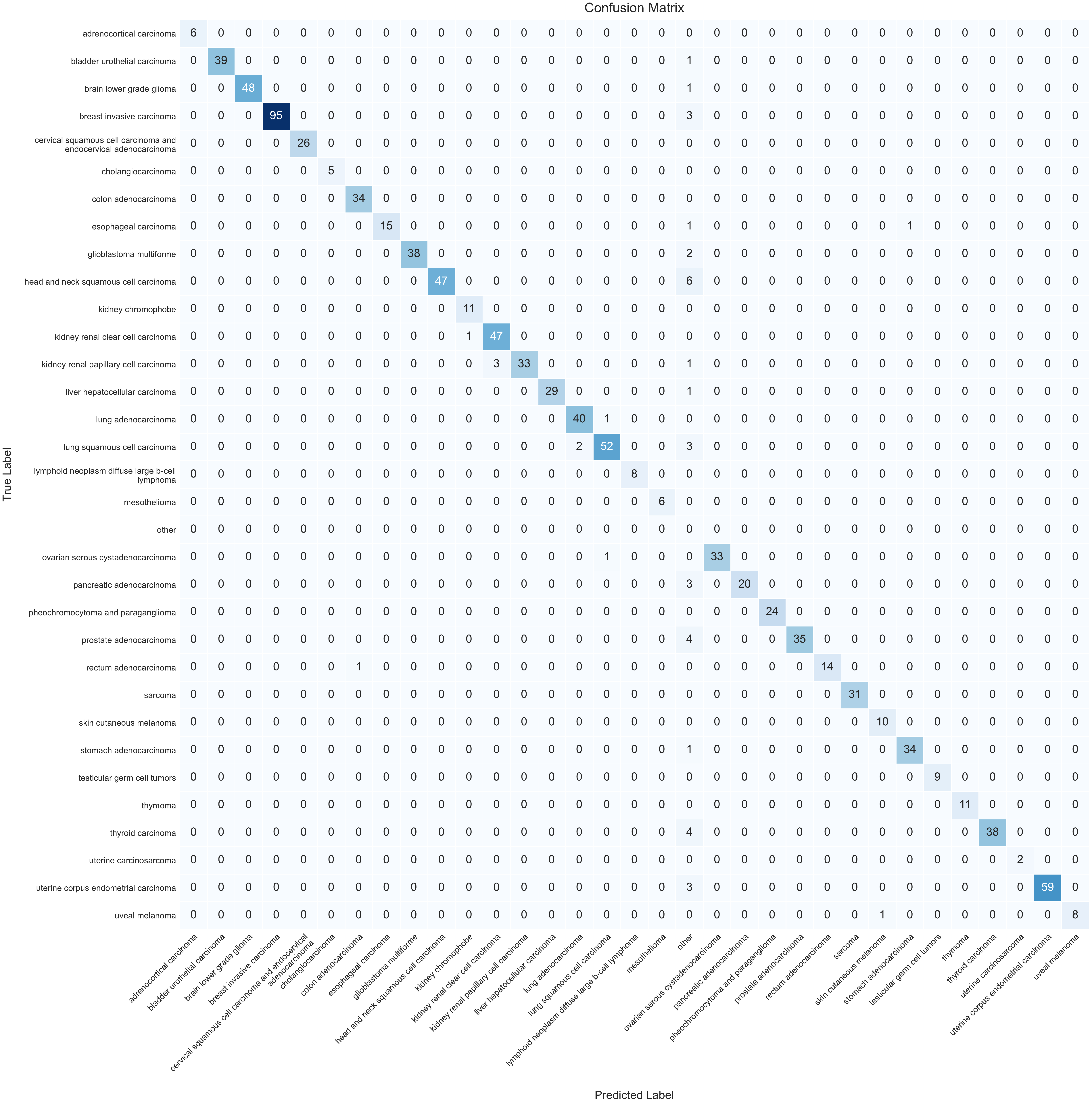}  
      \caption{Cancer Type Identification Confusion Matrix for Path-llama3.1-8B}
    \label{Task1-CM-PathLlama}
\end{figure}

\newpage

\begin{figure}[H]
      \centering
      \includegraphics[width=0.7\linewidth]{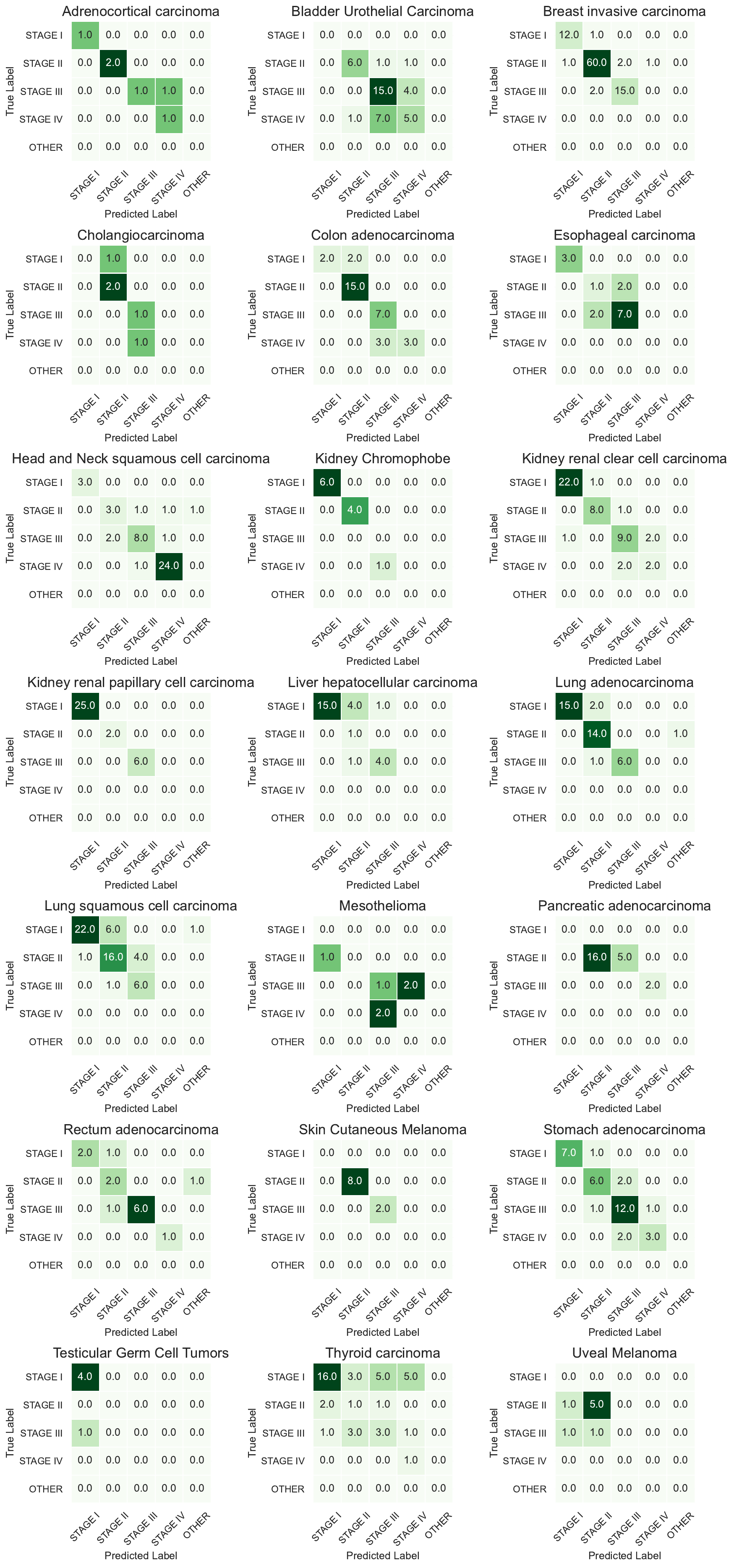}  
      \caption{AJCC Stage Identification Confusion Matrices for GPT-4o}
    \label{Task2-CM-GPT-4o}
\end{figure}

\newpage

\begin{figure}[H]
      \centering
      \includegraphics[width=0.7\linewidth]{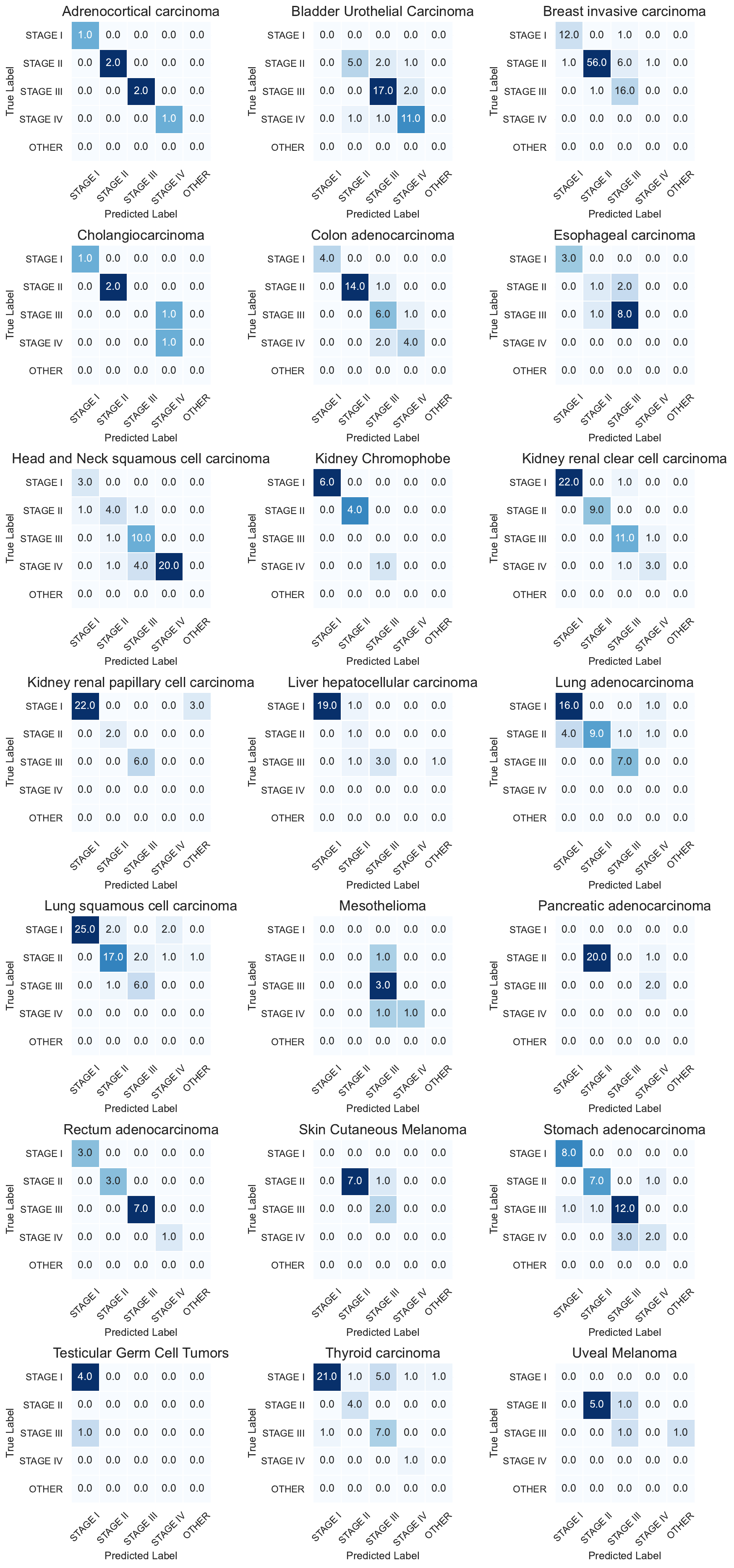}  
      \caption{AJCC Stage Identification Confusion Matrices for Path-llama3.1-8B}
    \label{Task2-CM-PathLlama}
\end{figure}

\newpage

\begin{figure}[H]
      \centering
      \includegraphics[width=0.7\linewidth]{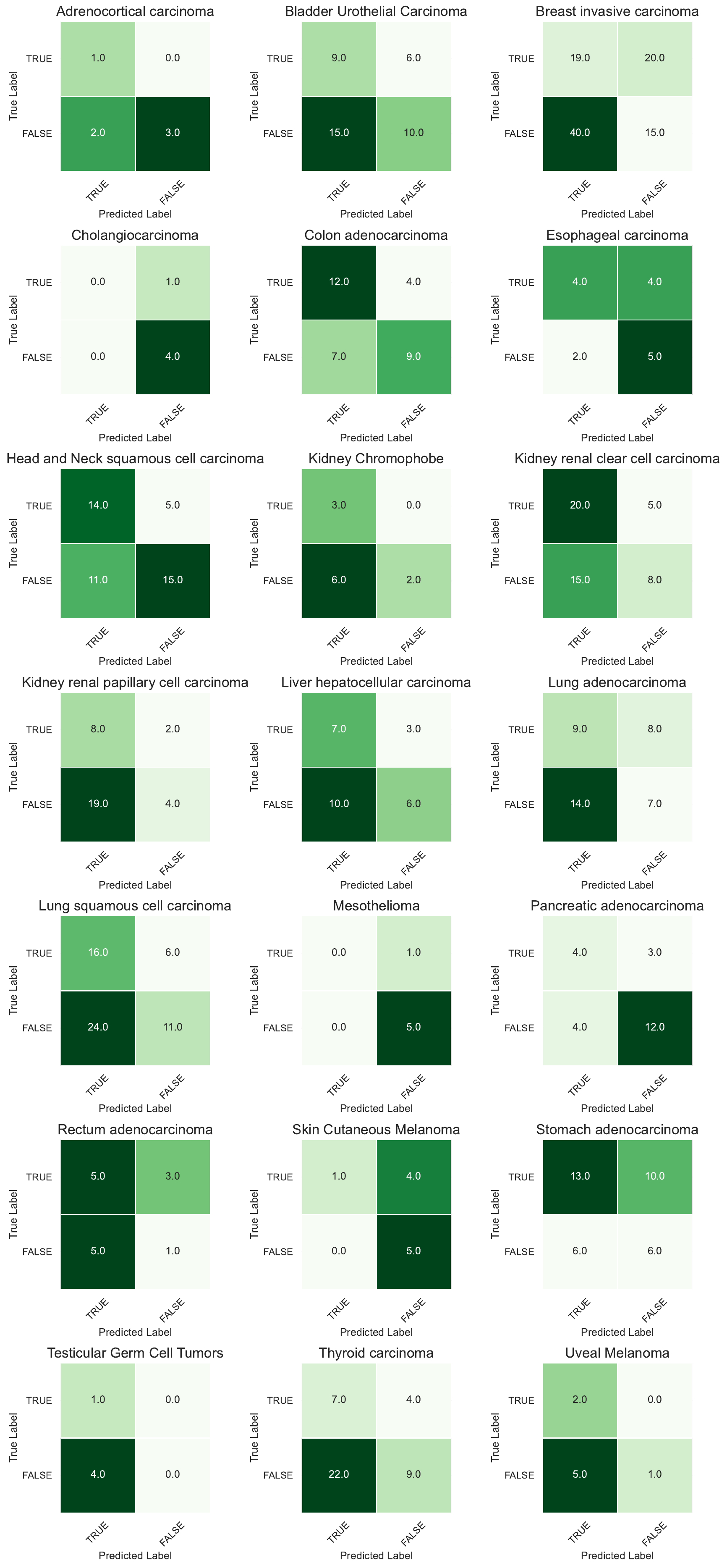}  
      \caption{Prognosis Assessment Confusion Matrices for GPT-4o}
    \label{Task3-CM-GPT-4o}
\end{figure}

\newpage

\begin{figure}[H]
      \centering
      \includegraphics[width=0.7\linewidth]{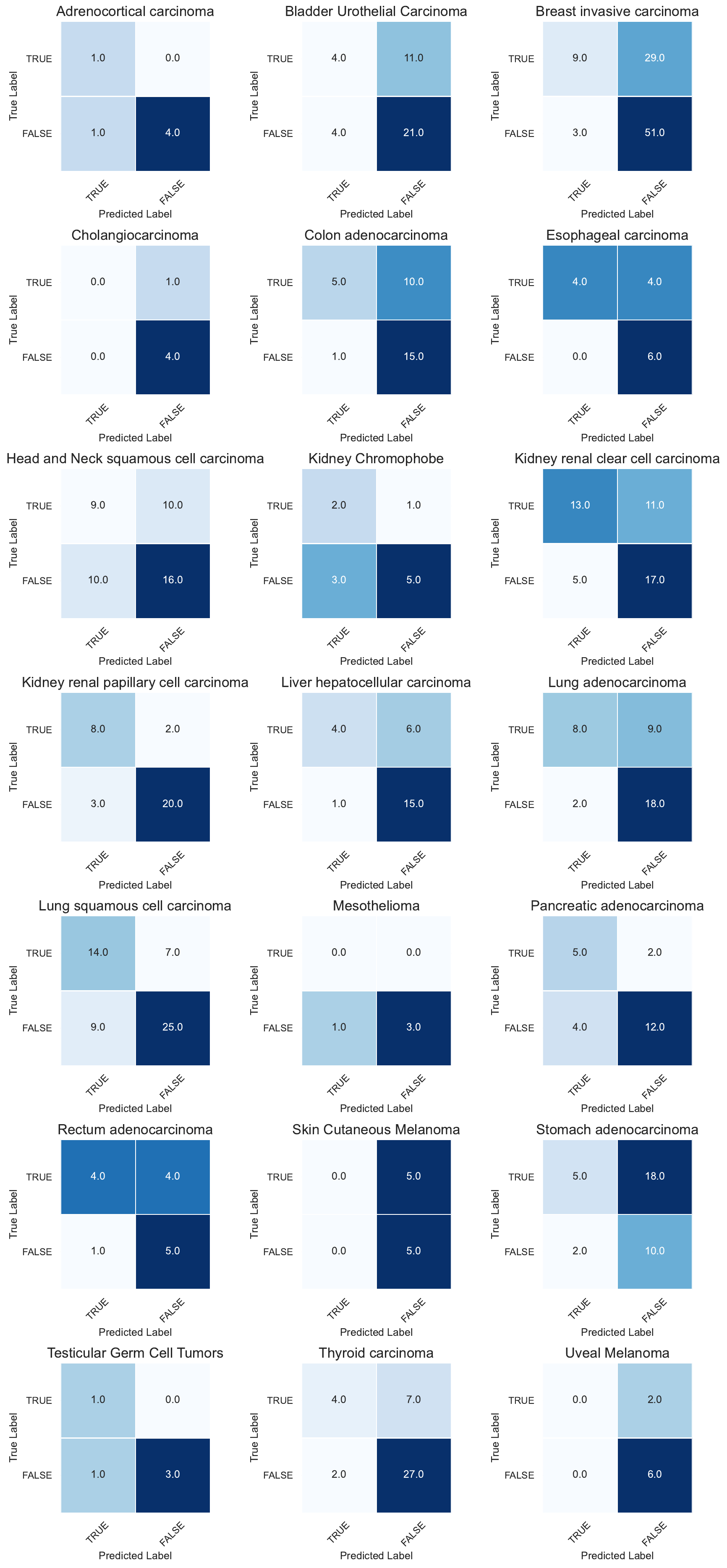}  
      \caption{Prognosis Assessment Confusion Matrices for Path-llama3.1-8B}
    \label{Task3-CM-PathLlama}
\end{figure}

\end{document}

%% file: 0_author_list.tex
\author{
Rachit Saluja\inst{1,2,3,*},
Jacob Rosenthal\inst{4}
\\
Yoav Artzi\inst{1,2},
David J. Pisapia\inst{3},
Benjamin L. Liechty\inst{3},
Mert R. Sabuncu\inst{1,2,3}
}

\authorrunning{Saluja, et al.}

\institute{\scriptsize{Cornell University, Ithaca, USA \and
Cornell Tech, New York, USA \and
Weill Cornell Medicine, New York, USA \and
Weill Cornell/Rockefeller/Sloan Kettering Tri-Institutional M.D.-Ph.D. Program, New York, USA
}
\\
\textsuperscript{*} Corresponding author: \email{\{rs2492@cornell.edu\}}}

%% file: 1_introduction.tex
\section{Introduction}

Cancer remains one of the leading causes of death globally \cite{siegel2023cancer}. The diagnosis and prognosis of cancer require the collaboration of numerous healthcare professionals, who invest considerable time and expertise to optimize patient outcomes. Despite this effort, current practices lack a cohesive, data-driven approach that could significantly enhance diagnosis and prognosis by leveraging the vast amounts of clinical information available in unstructured text within the medical record, such as pathology reports. Due to time and budget constraints, healthcare professionals are often unable to analyze and extract meaningful insights from the millions of pathology reports generated, which limits their ability to conduct comprehensive analyses for improving patient care.

Pathology reports, in particular, are complex and unstructured medical documents characterized by inconsistent formatting and a high degree of variability \cite{swillens2019identification, sluijter2016effects}. These reports often include extensive sections of information that are clinically irrelevant, making them less efficient for clinicians \cite{sluijter2016effects}. 
However, these reports also contain carefully curated information from highly trained domain-expert human pathologists in the form of diagnoses and descriptions of morphology. Extracting relevant data from pathology reports can provide valuable insights, revealing disease patterns, treatment outcomes, and overall patient health \cite{truhn2024extracting}. Utilizing this information can lead to more accurate prognosis predictions and the creation of personalized treatment plans \cite{dumbrava2018personalized, benary2023leveraging}. Beyond clinical use, this data is also essential for researchers constructing large-scale datasets and conducting medical studies.

Recent studies have demonstrated that large language models (LLMs) exhibit exceptional capabilities in leveraging their vast knowledge for tasks such as information extraction and certain forms of reasoning \cite{NEURIPS2020_1457c0d6,kojima2022large}. In the clinical domain, LLMs have shown remarkable potential, excelling at extracting data from medical texts without requiring extensive pretraining \cite{agrawal-etal-2022-large}. They have also proven highly effective in tasks like diagnosing complex medical cases \cite{doi:10.1056/AIp2300031, barile2024diagnostic, goh2024influence}, mining radiology reports for key insights \cite{fink2023potential, liu-etal-2023-exploring-boundaries, mukherjee2023feasibility, horiuchi2024accuracy}, and extracting information from electronic health records \cite{yang2022large, luo2022biogpt, guevara2024large}.

Despite these advances, the application of LLMs in natural language processing (NLP) tasks specific to pathology remains underexplored. A recent study employed a BERT-based model \cite{devlin-etal-2019-bert} for named entity recognition, focusing on extracting diagnostic elements from breast pathology reports. However, this approach was limited to a single specimen class and lacked generalizability to other organ systems and disease processes \cite{zeng2023improving}. More recently, large-scale LLMs like GPT have shown superior performance compared to supervised-trained models in various information extraction tasks \cite{sushil2024comparative}. Notably, GPT-4 was found to outperform supervised models in a zero-shot inference setting, without requiring additional training, although this study was also limited to breast cancer pathology reports.

In another study \cite{kefeli2024generalizable}, LLMs were used to estimate the TNM stage of cancers from pathology reports, a critical task for determining prognosis and guiding treatment decisions. However, cancer staging is not always systematically included in pathology reports, and previous methods for extracting TNM attributes relied on regular expressions or traditional NLP techniques. This study introduced a dedicated TNM staging LLM specifically designed for this task. Despite its utility, it does not provide comprehensive information about the anatomic extent of the cancer beyond tumor size, nodal involvement, and metastasis. As a result, relying solely on TNM staging may not fully capture the complexity of disease progression.

Moreover, these models are typically designed to handle a single task. Although they can be deployed locally, this setup may become expensive, as separate models are required to determine the tumor, nodes, and metastasis stages, respectively. An evaluation of the same task using open-source LLMs, combined with empirical tweaks to prompting techniques, assessed their performance in TNM staging \cite{change2024classifying}. However, a key limitation of these studies is that they tested their methodologies on extremely large models that require multiple GPUs with substantial VRAM, which is impractical for large-scale data mining, especially in resource-constrained settings. Furthermore, token generation with these large models is notably slow, further hindering their widespread application.

To address the issues mentioned above, we first evaluate the capabilities of both large-scale closed-source LLMs and open-source LLMs in performing information extraction tasks, such as identifying cancer types. Next, we assess their ability to holistically deduce the severity of cancer from pathology reports and predict the AJCC cancer stage. This staging system provides a more comprehensive assessment than the TNM stage, as it focuses beyond the anatomic extent of the cancer. Finally, we benchmark the performance of these LLMs in predicting patient prognosis from pathology reports alone, as a measure of performance on a complex reasoning task which requires both concrete information extraction and high-level reasoning.

After evaluating the performance of existing models, we developed a methodology to instruction-tune two new models that perform equally well, if not better, than both large-scale closed-source and open-source LLMs across all three tasks. The primary motivation for instruction tuning was to create a smaller, more efficient model capable of delivering high performance while being deployable in resource-constrained environments. We instruction-tuned GPT-4o-mini using a black-box fine-tuning approach provided by the OpenAI Platform, and Llama3.1-instruct-8B, which can be trained in an open-source environment. Notably, the Llama3.1-instruct-8B model, which we refer to as \textbf{Path-llama3.1-8B}, is a low-resource LLM capable of handling all three tasks efficiently and can be deployed in resource-constrained settings.

The key contributions of our study are as follows:
\begin{enumerate}
    \item We conduct a comprehensive evaluation of both open-source (Llama3-Instruct 8B and Llama3-Instruct 70B) \cite{dubey2024llama} and closed-source LLMs (GPT-4o, GPT-4o-mini, Mistral-Large, Mistral-Medium) \cite{hurst2024gpt, jiang2024mixtral} in zero-shot settings, focusing on information extraction tasks such as cancer type identification, as well as reasoning tasks like estimating the AJCC stage and predicting patient prognosis from pathology reports.
    \item To enhance the reliability and accuracy of predictions, we developed our own instruction-tuned models that perform as well as, or better than, existing models. Notably, one of these is an open-source model that can deliver accurate predictions even in low-resource environments.
    \item We open-source all code and trained models, making them easily deployable. Additionally, we provide the code required for further instruction tuning to adapt to hospital-specific scenarios, ensuring privacy and adaptability.
\end{enumerate}

%% file: 2_results.tex
\section{Results}

To evaluate the performance of LLMs in cancer type identification, AJCC stage determination, and prognosis assessment, we begin by curating a custom dataset, the specifics of which are outlined in Section \ref{Section-Methods}. The dataset is structured into three question-and-answer tasks, each addressing a specific aspect: cancer type identification, stage classification, and prognosis assessment. In each task, the question combines a query related to the respective topic—cancer type, stage, or prognosis—with the corresponding pathology report, and the answer represents the desired outcome. Given the verbosity of chat-based LLMs, we constrain their outputs to strictly adhere to a JSON object format \cite{liu2024we}, structured as key-value pairs. This approach facilitates answer extraction using regular expressions for evaluation purposes while ensuring the outputs are well-suited for subsequent data mining and information extraction tasks.

For each task, we evaluate the performance of six models: two OpenAI models (GPT-4o and GPT-4o-mini)\cite{hurst2024gpt}, two Mistral models (Mistral-Medium and Mistral-Large) \cite{jiang2024mixtral}, and two Llama 3 models (the 8B and 70B variants)\cite{dubey2024llama}. Each experiment is conducted five times per model to ensure robust interpretations and to compute reliable confidence intervals. We perform instruction tuning by generating synthetic question-answer pairs, which, when combined with pathology reports, form a diverse and comprehensive training dataset. The\textbf{ Path-GPT-4o-mini-FT} model is fine-tuned using OpenAI’s proprietary platform, while \textbf{Path-llama3.1-8B} undergoes fine-tuning via Low-Rank Adaptation (LoRA) \cite{hu2022lora}, specifically optimizing key attention components. More details on data curation and model training are provided in Section \ref{Section-Methods}.

\begin{figure}[H]
      \centering
      \includegraphics[width=1\linewidth]{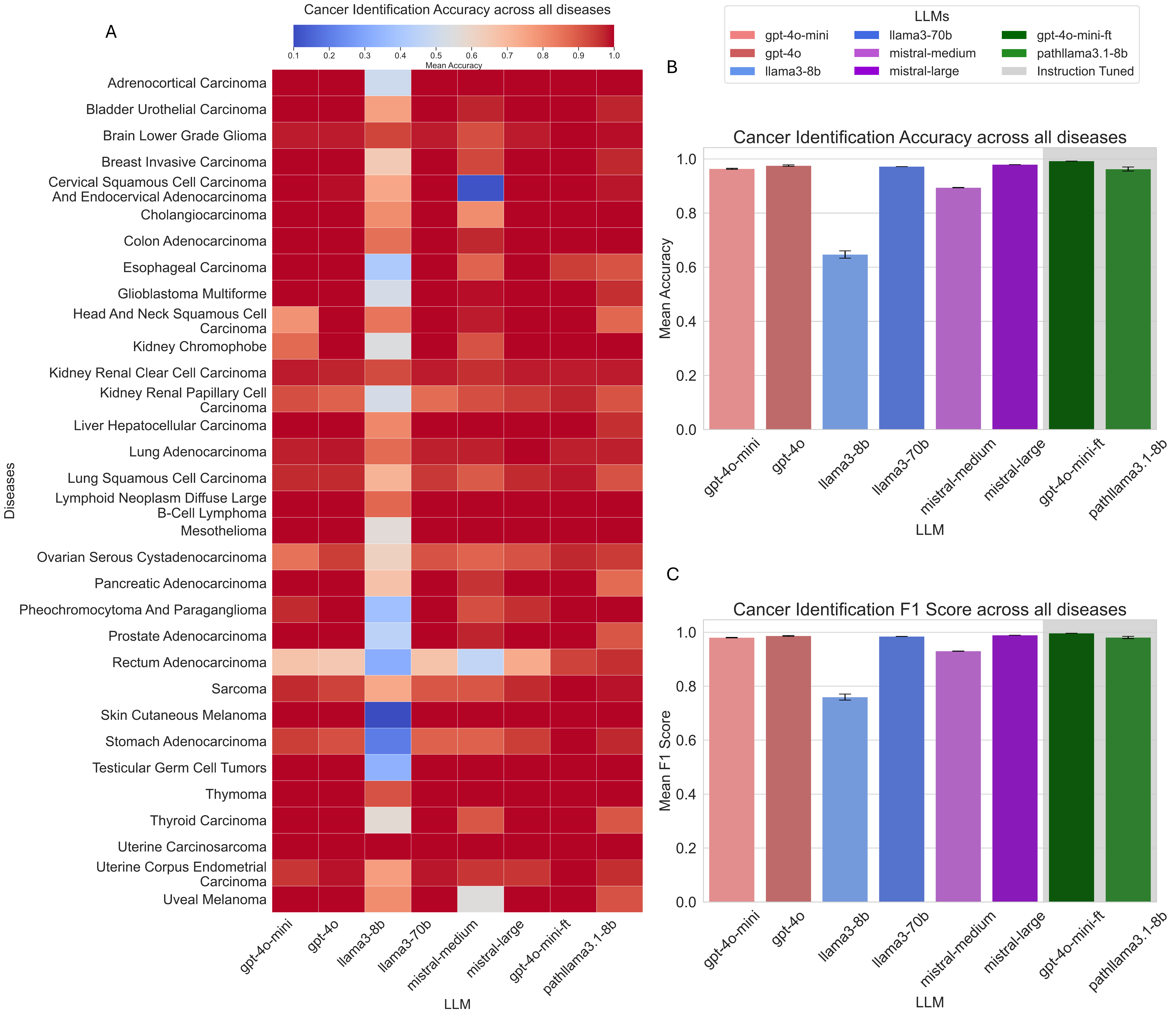}  
      \caption{
      (A) Heatmap illustrating cancer identification accuracy across all diseases. 
    (B)  Cancer identification accuracy metrics.
    (C)  Cancer identification F1-score metrics.
      } 
    \label{Fig1-1-Performance}
\end{figure}

\subsection{Cancer Type Identification}

The six language models and two instruction-tuned models were evaluated on a test set comprising 952 pathology reports. These reports represent 32 distinct cancer types from various human body systems, providing a diverse and comprehensive dataset for evaluation. All eight models were assessed under identical system conditions and standardized user prompts, with outputs strictly formatted as JSON objects. The evaluation followed a zero-shot answering mechanism, as illustrated in Figure \ref{Fig4-Framework}(A), with further details of the prompts provided in Appendix \ref{Appendix-Prompts}.

As depicted in Figure \ref{Fig1-1-Performance}, most language models demonstrated strong performance in cancer type identification, achieving a mean accuracy exceeding 96\%, with the exceptions of Llama3-8B and Mistral-Medium. While Mistral-Medium achieved an accuracy of 90\%, the Llama3-8B model performed significantly worse, with a mean accuracy of 64\%. This underperformance is largely attributed to the model's smaller parameter size and limited capability to follow instructions in tasks requiring information extraction from pathology text reports.

\begin{figure}[H]
      \centering
      \includegraphics[width=1\linewidth]{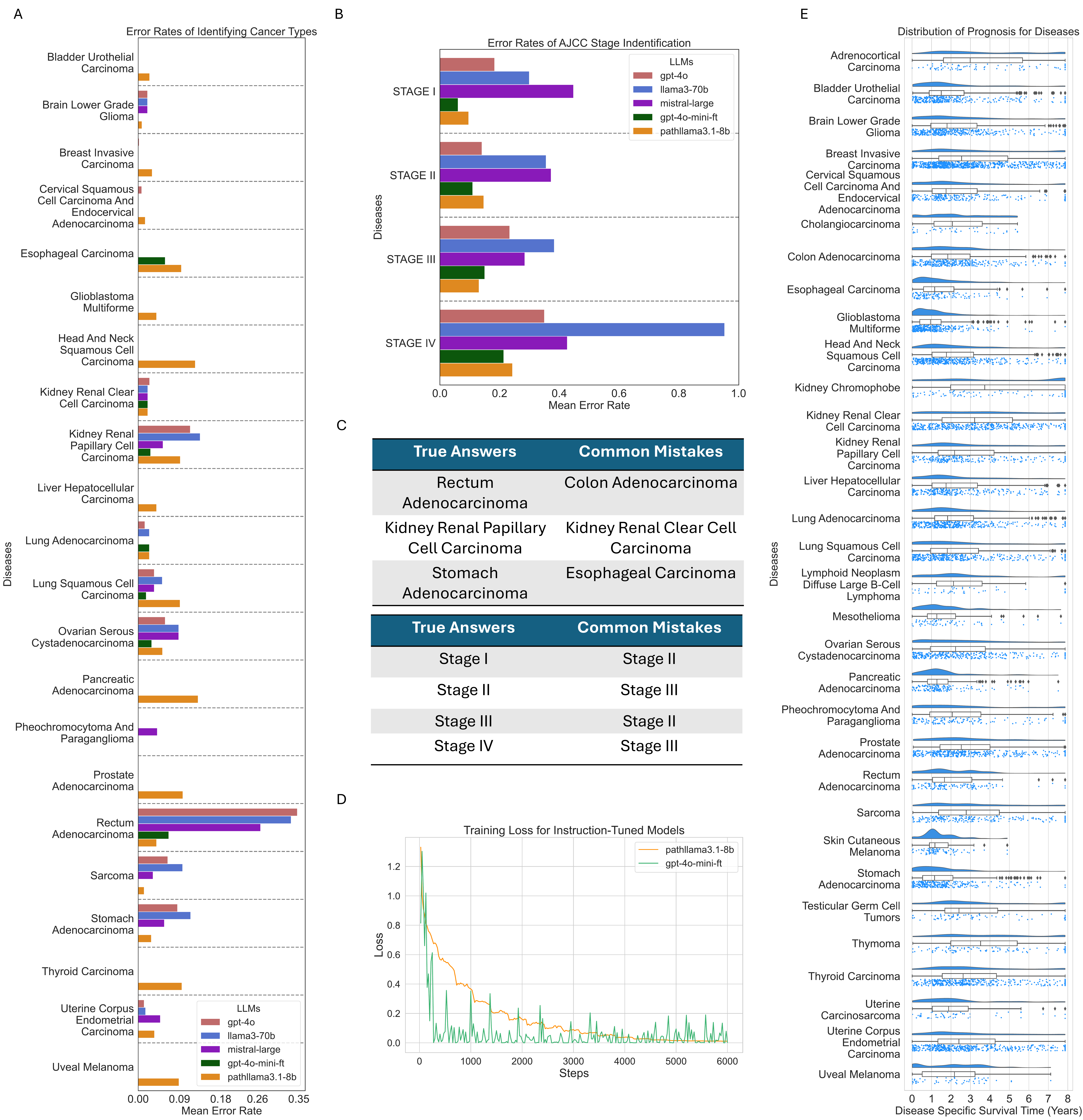}  
      \caption{
    (A) Error rates of identifying cancer types.
    (B) Error rates of identifying AJCC Stage.
    (C) Common mistakes made by LLMs.
    (D) Training loss curves for the instruction-tuned models, plotted only for 6,000 steps, although GPT-4o-mini was instruction-tuned for 17,000 steps.
    (E) Distribution of Disease-Specific Survival Time in years.
      }
    \label{Fig2-Error-rates}
\end{figure}


The fine-tuned GPT-4o-mini model, referred to as \textbf{Path-GPT-4o-mini-FT}, outperformed all other large language models, achieving a mean accuracy of 99\%. This improvement is credited to its training on the instruction-tuning dataset, which not only enhanced its accuracy but also ensured consistent adherence to the JSON object output format. Similarly, the instruction-tuned Llama3.1 model, named \textbf{Path-llama3.1-8B}, exhibited a notable performance boost when fine-tuned on the same dataset, achieving a mean accuracy of 96\%. This marks a substantial improvement for a model with only 8 billion parameters, highlighting the effectiveness of the fine-tuning process.

Cancer type identification, being primarily an information extraction task, is the easiest of the three tasks. All LLMs exhibit high performance, with the exception of Llama3-8B, which struggles due to its smaller capacity and limited instruction-following capabilities. While Llama3-8B may successfully extract the correct cancer type, it fails to meet our strict evaluation criteria, designed with software deployment in mind, as it often produces outputs that deviate from the required format. However, with instruction tuning, the same-sized model, \textbf{Path-llama3.1-8B}, performs significantly better. This improvement is because the model learns to produce the exact output format required, making it more suitable for deployment in downstream tasks.

The other models perform well during inference, likely due to their robust pretraining on extensive and diverse datasets, which may include pathology-related text. Furthermore, these models have undergone instruction tuning through various chat-tuning frameworks that optimize their response generation using human feedback \cite{rafailov2024direct, ouyang2022training}. This fine-tuning process significantly enhances their ability to provide accurate, context-aware, and user-aligned responses, including outputs formatted as JSON objects, as required in this task.

From Figure \ref{Fig2-Error-rates}, it is evident that the errors made by the LLMs are relatively straightforward and follow similar patterns across models. 
Figure \ref{Fig2-Error-rates}(A) highlights the most common diseases and their corresponding error rates for both larger models and instruction-tuned models. A key observation is that all models tend to exhibit similar behavior when making mistakes, often confusing anatomically or functionally related organs, leading to incorrect predictions. For example, as shown in Figure \ref{Fig2-Error-rates}(C), Rectum Adenocarcinoma is frequently misclassified as Colon Adenocarcinoma\footnote{The distinction between colon and rectal adenocarcinoma remains debated, as they are frequently classified together under the term \textbf{colorectal adenocarcinoma}.}. Similarly, confusion arises between Stomach Adenocarcinoma and Esophageal Carcinoma, reflecting the challenge of distinguishing closely related organ-specific cancers. Another notable misclassification occurs between Kidney Renal Papillary Cell Carcinoma and Kidney Renal Clear Cell Carcinoma. This further highlights how LLMs can struggle to distinguish between different cell types, leading to misclassifications. However, instruction tuning significantly reduces these error rates, demonstrating its effectiveness in improving the models' ability to handle subtle distinctions and deliver more accurate predictions.

\subsection{AJCC Stage Identification}

The same set of models was also evaluated on a dataset of 594 pathology reports to determine their corresponding stage (Stage I, II, III, or IV). These reports represent 21 distinct cancer types. The evaluation conditions were identical to those used for cancer type identification, with one key modification: a self-generated chain-of-thought query was included in the system prompt to guide the model through a step-by-step reasoning process to arrive at a specific answer. While this approach increased both the input and output token counts, it significantly improved performance. However, the instruction-tuned models did not utilize the chain-of-thought methodology and instead adhered to a zero-shot answering mechanism.

\begin{figure}[H]
      \centering
      \includegraphics[width=1\linewidth]{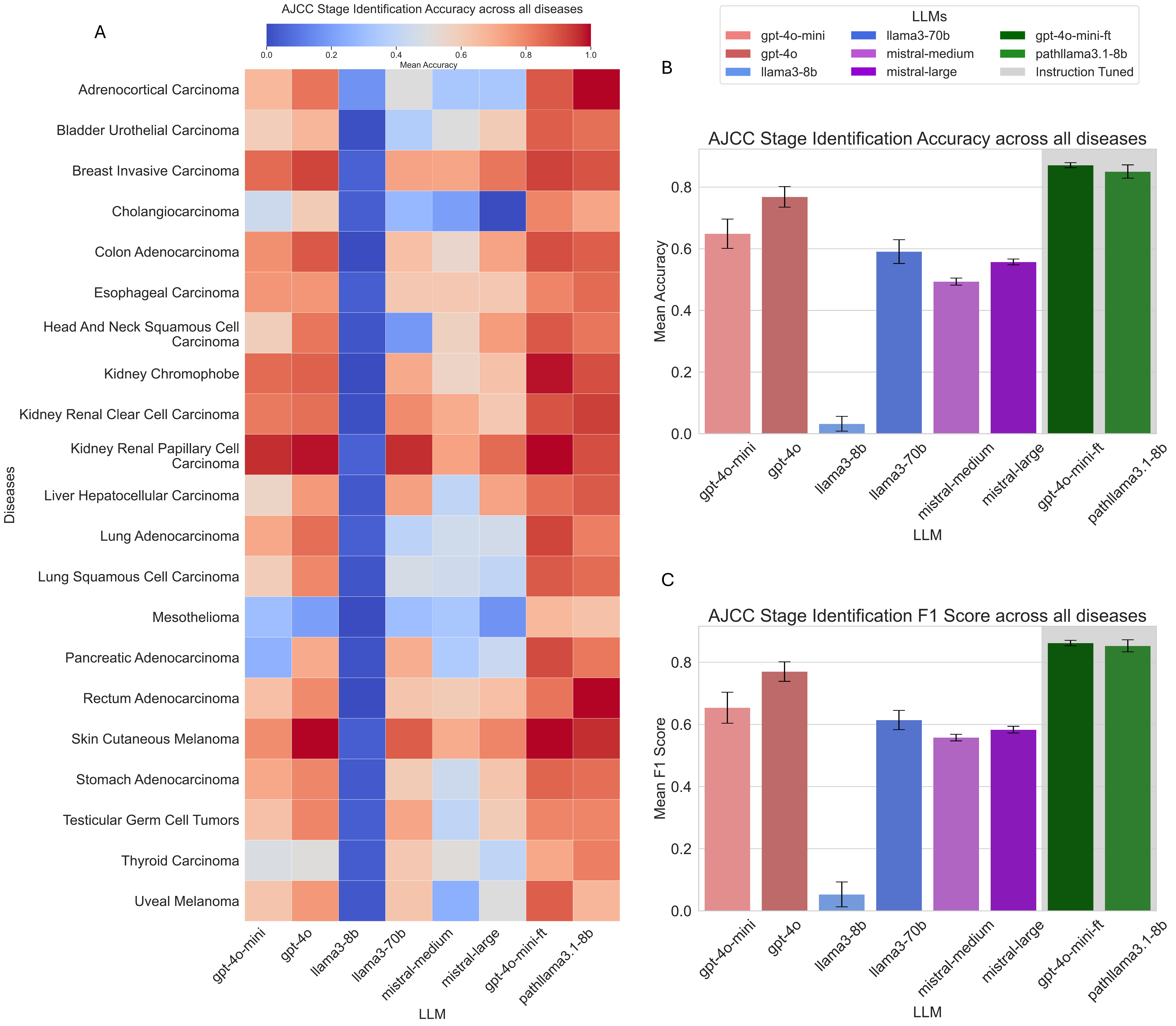}  
      \caption{
      (A)  Heatmap illustrating AJCC stage identification accuracy
across all diseases.
    (B)  AJCC stage identification accuracy metrics.
    (C)  AJCC stage identification F1-score metrics.
      } 
    \label{Fig1-2-Performance}
\end{figure}

As illustrated in Figure \ref{Fig1-2-Performance}, the performance of all LLMs is notably lower for AJCC stage identification compared to their results for cancer type identification. This discrepancy arises because stage estimation involves deductive reasoning, requiring not only insights from the provided text but also some level of world knowledge to make an informed prediction. A particularly notable outcome is the significant drop in performance of the Mistral models, which had performed well in the previous task. Similarly, the Llama3-8B model performs poorly, often deviating from the expected answer by producing irrelevant outputs, such as TNM staging information. Among the non-instruction-tuned models, OpenAI's GPT-4o achieves the highest mean accuracy, reaching 76\%.

The instruction-tuned models, however, demonstrate remarkable performance compared to running standard inference with LLMs. The \textbf{Path-GPT-4o-mini-FT} model achieves a mean accuracy of 87\%, while the \textbf{Path-llama3.1-8B} model attains a mean accuracy of 85\%. Notably, these results are achieved without employing a chain-of-thought approach. It is particularly significant that the \textbf{Path-llama3.1-8B}, despite being an 8B model, outperforms GPT-4o on this task. This underscores the potential of smaller, fine-tuned models to achieve high performance on specialized tasks, even surpassing larger, closed-source LLMs in certain scenarios.

Determining the AJCC stage is inherently more challenging. In our initial experiments, we observed that when using a zero-shot approach, LLMs often default to predicting either Stage I or Stage IV. We speculate that this behavior stems from the training data these models are exposed to, which predominantly consists of internet text. Such data often lacks detailed clinical specifics and tends to portray cancer as a life-threatening condition, leading the models to overpredict Stage IV. Conversely, predicting Stage I may serve as a precautionary default response in the absence of sufficient context. In another experiment, we explored a two-step zero-shot prompting mechanism. In this approach, the LLMs were first asked to output the TNM stage, followed by determining the AJCC stage based on the provided TNM information. However, this method proved ineffective due to excessive hallucinations by the LLMs during the initial step, where the models often generated inaccurate or inconsistent TNM stage information, ultimately compromising the reliability of the subsequent AJCC stage prediction.

To address the limitations observed in earlier experiments, we next explored an empirical approach that encouraged LLMs to implicitly reason and leverage their world knowledge. This was achieved by employing self-generated chain-of-thought (CoT) reasoning. As shown in Figure \ref{Fig3-Cot-Trace}, an LLM successfully creates a reasoning trace for Skin Cutaneous Melanoma that aligns closely with the AJCC staging guidelines. For instance, GPT-4o systematically extracts tumor size features, evaluates the spread of cancer to lymph nodes, and examines the presence of distant metastases. Moreover, it accurately identifies ulceration, a critical feature that can elevate the AJCC stage classification. In this particular example (Figure \ref{Fig3-Cot-Trace}), the model avoids hallucination, refraining from discussing features like the mitotic rate, which are absent from the report. This demonstrates the potential of CoT reasoning to improve the reliability and accuracy of complex staging tasks.

\vspace{-0.5cm}

\definecolor{RoyalGray}{RGB}{66, 66, 54}

\begin{figure}[H]
\centering
\begin{tcolorbox}[
    colback=black!5!white,    
    colframe=RoyalGray,      
    title=Chain of thought trace for Skin Cutaneous Melanoma (AJCC Stage Identification),
    fonttitle=\bfseries,  
    coltitle=white,          
    colbacktitle=RoyalGray,  
    boxrule=1mm,             
    width=\textwidth         
]

\textbf{Pathology Report:} Path Report:SKIN TISSUE CHECKLIST. Specimen type: Excision of tumor. Tumor site: Skin. Tumor size: 3 x3x1.5cm. Tumor features: Ulcerated. Satellite nodules: Not specified. Histologic type: Nodular melanoma. Histologic grade: Lymph nodes: Not specified. Lymphatic invasion: Not specified. Venous invasion: Not specified. Margins: Not specified. Evidence of neo-adjuvant treatment: Not specified. Additional pathologic findings: Not specified. Comments: 15 mm deep, III Level by Clarc.

\vspace{0.1cm}

\textbf{GPT-4o Output:} The pathology report provides the following key information:

\vspace{-0.3cm}

\begin{enumerate}
    \item Tumor Size: The tumor measures 3 x 3 x 1.5 cm.
    \item Tumor Features: The tumor is ulcerated.
    \item Histologic Type: Nodular melanoma.
    \item Depth: 15 mm deep, which corresponds to a Clark Level III.
\end{enumerate}

\vspace{-0.3cm}

To determine the AJCC stage, we need to consider the depth of invasion, ulceration, and any lymph node involvement or metastasis, although the latter two are not specified in the report.

\vspace{-0.3cm}

\begin{enumerate}
    \item Depth of Invasion: A depth of 15 mm is significant. In melanoma staging, depth (Breslow thickness) is a critical factor. A depth greater than 4 mm is typically associated with higher stages.
    \item Ulceration: The presence of ulceration is a poor prognostic factor and can upstage the melanoma.
    \item Clark Level III: This indicates invasion into the papillary dermis, but the depth (15 mm) is more critical for staging.
\end{enumerate}

\vspace{-0.3cm}

Given the depth of 15 mm and the presence of ulceration, this melanoma is likely to be at least Stage II. However, without information on lymph node involvement or distant metastasis, we cannot definitively classify it as Stage III or IV.
Therefore, based on the provided information, the most appropriate stage is Stage II, as it accounts for the significant depth and ulceration without evidence of further spread. \newline
Final Answer:
\{"stage": "Stage II"\}

\

\setlength{\tabcolsep}{4pt} 

\begin{tabularx}{\textwidth}{X|>{\hspace{5pt}}X} 
\uline{\textbf{AJCC Staging Guidelines}}: & \uline{\textbf{Chain of thought trace (GPT-4o)}}: \\ 
T (Primary Tumor) & Able to understand dimensions of the tumor  \includegraphics[height=1em]{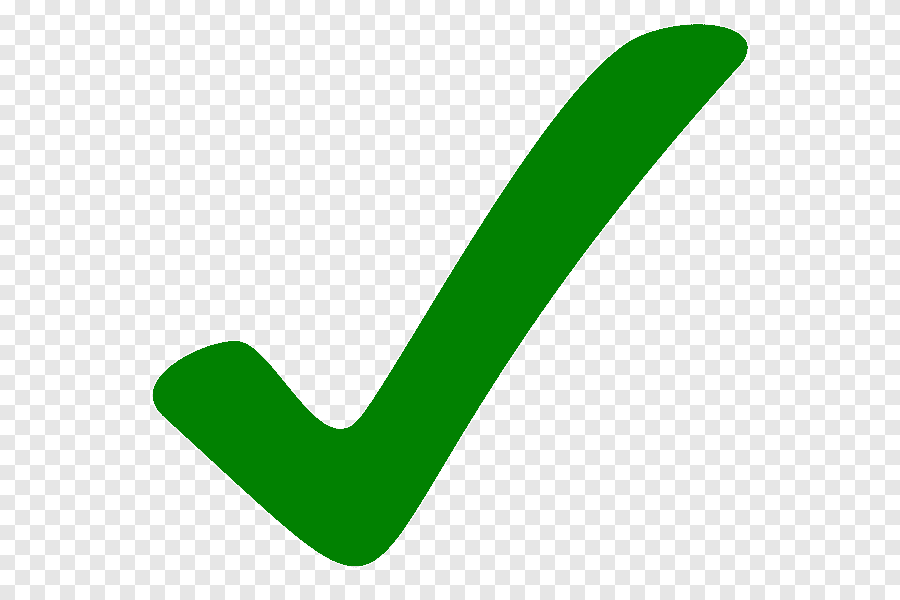} \\
Ulceration & Identifies presence of ulceration \includegraphics[height=1em]{extras/correct_symbol.png} \\
Mitotic Rate & Abstains from speaking about Mitotic Rate \\
N (Regional Lymph Nodes) & Identifies no lymph node involvement \includegraphics[height=1em]{extras/correct_symbol.png} \\
M (Distant Metastasis) & Identifies no distant metastasis \includegraphics[height=1em]{extras/correct_symbol.png}\\
\end{tabularx}
\end{tcolorbox}
\caption{Example of how GPT-4o performs using Chain of thought in comparison to the standard AJCC Guidelines for Skin Cutaneous Melanoma}
\label{Fig3-Cot-Trace}
\end{figure}

\vspace{-0.5cm}

While chain-of-thought (CoT) reasoning certainly enhances performance and holds considerable potential for zero-shot inference, our instruction-tuned models demonstrate significantly better results. Instruction tuning aligns the model's internal representations to interpret what a specific stage signifies in the context of a pathology report. This targeted alignment enables even an 8B parameter model, which requires minimal hardware\footnote{Can be trained on a NVIDIA 40x Series GPUs.} for execution, to achieve high accuracy and F1 scores, highlighting the efficiency and practicality of instruction-tuned models for such tasks.

The mistakes observed in this task are particularly intriguing. Across all models, the average error rates are fairly similar, with the exception of Stage IV, where the error rate is slightly higher. A common pattern in the errors is that stages are often mispredicted as being one stage higher or lower than the correct stage. This suggests that while the models grasp the general progression of staging, they sometimes struggle with the finer distinctions required to accurately assign the correct AJCC stage.

From Figure \ref{Fig2-Error-rates}(B) and (C), we observe that Stages II and III are often confused with one another, likely due to ambiguous interpretations of lymph node involvement in relation to the cancer. While Stage IV has strict guidelines requiring the presence of distant metastases for classification, it is notable that many instances are incorrectly classified as Stage III. This discrepancy warrants further investigation to better understand the reasoning behind such errors and to identify potential gaps in the models’ comprehension of AJCC staging criteria.

\begin{figure}[H]
      \centering
      \includegraphics[width=1\linewidth]{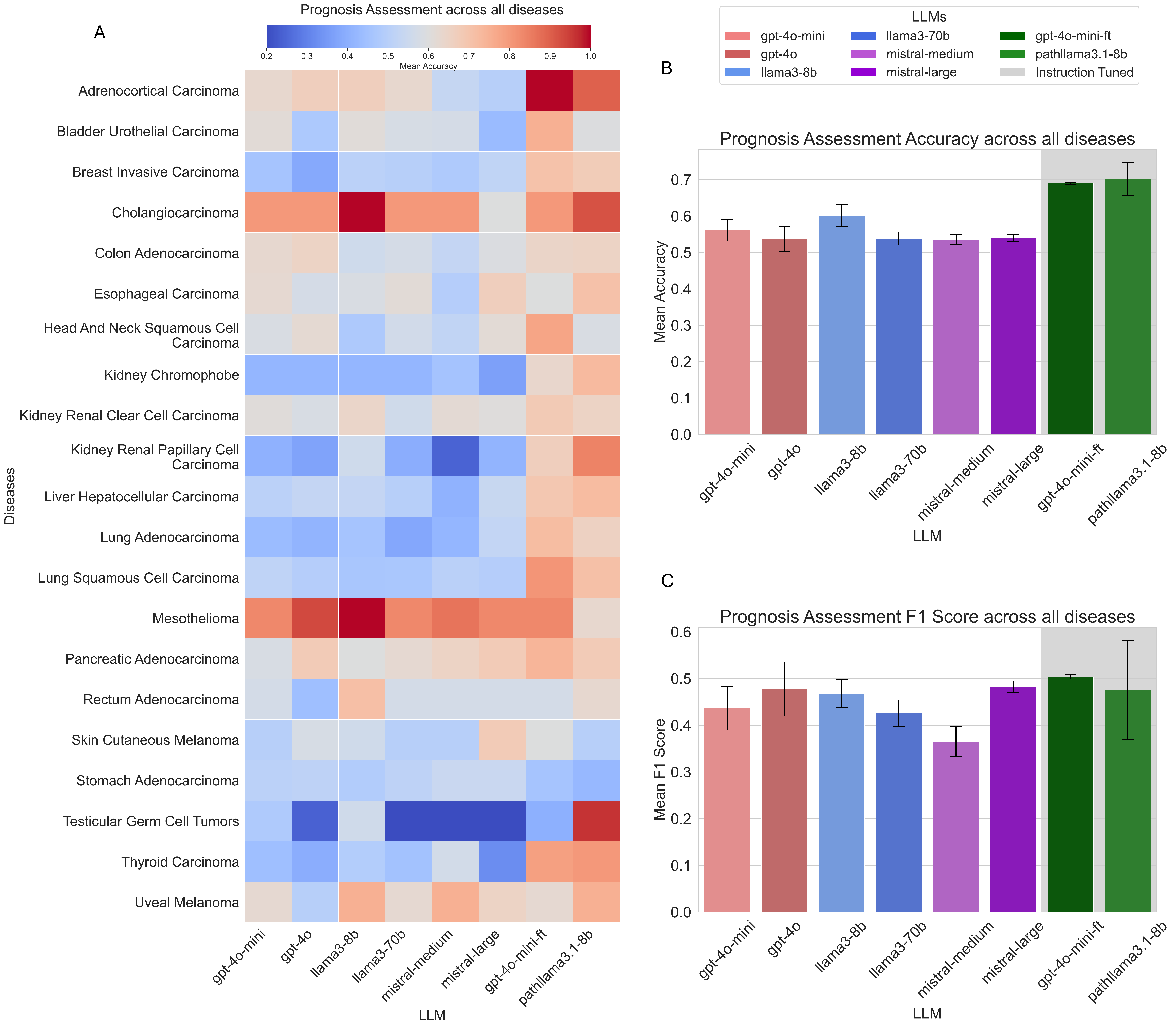}  
      \caption{
      (A) Heatmap illustrating prognosis assessment accuracy across all diseases.
    (B)  Prognosis assessment accuracy metrics.
    (C)  Prognosis assessment F1-score metrics.
      } 
    \label{Fig1-3-Performance}
\end{figure}

\subsection{Prognosis Assessment}

For prognosis assessment, the task is formulated as a binary classification problem. The task is to predict whether a patient will survive beyond the disease-specific \textit{mean survival time} based on the input pathology report. LLMs are provided with 8 examples (i.e., 8 - consisting of 4 positive and 4 negative instances. Additionally, a self-generated chain-of-thought query is incorporated into the system prompt to guide the reasoning process. 

The models were evaluated on 593 pathology reports to assess prognosis, with the dataset encompassing 21 distinct cancer types. Among the three tasks, prognosis assessment proves to be the most challenging, as it involves a two-step process requiring both deductive reasoning and future outcome prediction. Notably, the Llama3-8B model achieves the highest accuracy; however, for this task, the F1 score provides a more comprehensive evaluation metric due to binary class imbalance. GPT-4o and Mistral-Large demonstrate higher F1 scores, primarily due to their better handling of the minority class in the prediction problem. Despite these observations, the task remains inherently difficult, as most models struggled to consistently produce correct answers.

Once again, the instruction-tuned models demonstrate superior performance, achieving both higher accuracy and higher F1 scores. Notably, the \textbf{Path-GPT-4o-mini-FT} model achieves the highest accuracy as well as the highest mean F1 score. It is particularly significant that this performance is attained without employing chain-of-thought reasoning or the 8-shot approach, highlighting the effectiveness of instruction tuning in enhancing model performance for this complex task.

\vspace{-0.5cm}

\definecolor{RoyalGray}{RGB}{66, 66, 54}

\begin{figure}[H]
\centering
\begin{tcolorbox}[
    colback=black!5!white,    
    colframe=RoyalGray,      
    title=Chain of thought trace for Stomach Adenocarcinoma (Prognosis Assessment),
    fonttitle=\bfseries,  
    coltitle=white,          
    colbacktitle=RoyalGray,  
    boxrule=1mm,             
    width=\textwidth         
]

\textbf{Pathology Report:} Subiect ID. Final Pathology Template. Final Pathology Report Information (Surgical Tissue Case). 1.1. Tissue Specimen ID Code: 1.2. Date of Excision: (MM/DD/YYY). 1.3. Tumor Location (anatomic site): Stomach, Antrum. 1.4. Tumor Size: 3,0x 4,0 (cm). 1.5. Lymph Node Status: 1.5.1. Number Nodes Examined: 11. 1.5.2. Number Nodes Positive: 0. 1.5.3. Lymph Node Locations (Involved Nodes): 8 greater curvature,. 3. lesser curvature. 1.6. Metastasis: 1.6.1. Metastasis Present: Unknown. 1.6.2. Number of Mets: 1.6.3. Site(s) Of Metastasis: 1.7. ICD-9 Code for Specimen Diagnosis: C 16.3. 1.8. Final Pathology Details: (Please suppy de-identified copy of original language final pathology report in. addition to information below). 1.8.1 Macroscopic Description: Surgical specimen of stomach,. measuring 16 cm in combination with greater omentum. At. 1,0 cm from the resection margin, ulcerative formation. with irregular borders 3.0 x 4.0 cm in dimension,. macroscopically not infiltrating gastric wall. Greater. omentum is intact. 11 lymph nodes (8 - greater. curvatures, 3 - lessen curvature). 1.8.2 Microscopic Description: Adenocarcinoma diffuse type. undifferentiated of stomach, with signet ring cell type. regions. Tumor invades through muscularis propria. Margins resection is intact. Greater omentum is intact. Lymph node status negative: greater curvature (8 from 8). and lesser curvature (3 form 3). 1.8.3. Comment (if applicable): 1.8.4. Clinical Diagnosis: Gastric cancer of antrum. 1.8.5. Stage: T2a NO M0 (TxNxMx). hu8819913 Yes. PROPRIETARY AND CONFIDENTIAL.

\vspace{0.1cm}

\textbf{GPT-4o Output:} To determine the likelihood of survival after 1.54 years based on the provided pathology report, let's analyze the key details:

\vspace{-0.3cm}

\begin{enumerate}
    \item Tumor Type and Differentiation: The tumor is an undifferentiated diffuse type adenocarcinoma with signet ring cell features. This type of cancer is generally more aggressive and has a poorer prognosis compared to other types of gastric adenocarcinomas.
    \item Tumor Size and Location: The tumor is located in the antrum of the stomach and measures 3.0 x 4.0 cm. The size is moderate, but the location in the antrum is common for gastric cancers.
    \item Lymph Node Involvement: There is no lymph node involvement, as all 11 examined nodes were negative for cancer (N0). This is a favorable prognostic factor, as lymph node involvement often indicates a higher risk of metastasis and poorer outcomes.
    \item Tumor Invasion: The tumor invades through the muscularis propria (T2a), which indicates that the cancer has penetrated into the muscle layer of the stomach wall but not beyond it. This is an early stage of invasion and is associated with a better prognosis compared to more advanced stages.
    \item Resection Margins and Greater Omentum:
    The resection margins and greater omentum are intact, meaning there is no residual tumor left after surgery. This is a positive factor for survival.
    \item Clinical Stage: The clinical stage is T2a N0 M0, indicating that the tumor is localized without regional lymph node involvement or distant metastasis. This stage is considered early and generally has a better prognosis.
\end{enumerate}

\vspace{-0.3cm}

Considering these factors:

\vspace{-0.3cm}

\begin{enumerate}
    \item The absence of lymph node involvement
    \item Early stage of invasion
    \item Clear resection margins
\end{enumerate}

\vspace{-0.3cm}

suggests a favorable prognosis. Despite the aggressive nature of undifferentiated diffuse type adenocarcinoma with signet ring cell features, the overall clinical picture indicates a reasonable likelihood of survival beyond 1.54 years. Answer: \{"Survival": "True"\}

\end{tcolorbox}
\caption{Example of how GPT-4o performs using Chain of thought for Prognosis Assessment}
\label{Fig4-Cot-Trace-Prog}
\end{figure}

\vspace{-0.5cm}

Analyzing the performance of LLMs for prognosis is the most challenging of the three tasks. Even framing the problem required considerable effort, as estimating the number of years a patient is expected to survive based on a pathology report inherently resembles a regression problem. However, LLMs are notoriously poor at such tasks due to their lack of statistical understanding and inability to effectively interpret the complex interplay of features required for accurate prognosis estimation. This limitation highlights the need for alternative approaches to better leverage LLM capabilities for this problem domain. 

To address the challenges, the problem was reframed by incorporating the disease-specific survival time into the dataset, providing the LLM with a virtual threshold to determine whether a patient would survive beyond that time. Additionally, we observed that relying solely on self-generated chain-of-thought reasoning was insufficient for achieving reliable results. To enhance performance, we provided a few example cases as references, enabling the model to better understand the task and produce more accurate answers. Despite these adjustments in methodology, most LLMs still fail to perform at a satisfactory level, indicating significant room for improvement. This underscores the inherent complexity of the prognosis task and highlights the limitations of current LLM capabilities in accurately addressing such nuanced problems.

\vspace{-0.5cm}

\begin{figure}[H]
      \centering
      \includegraphics[width=1\linewidth]{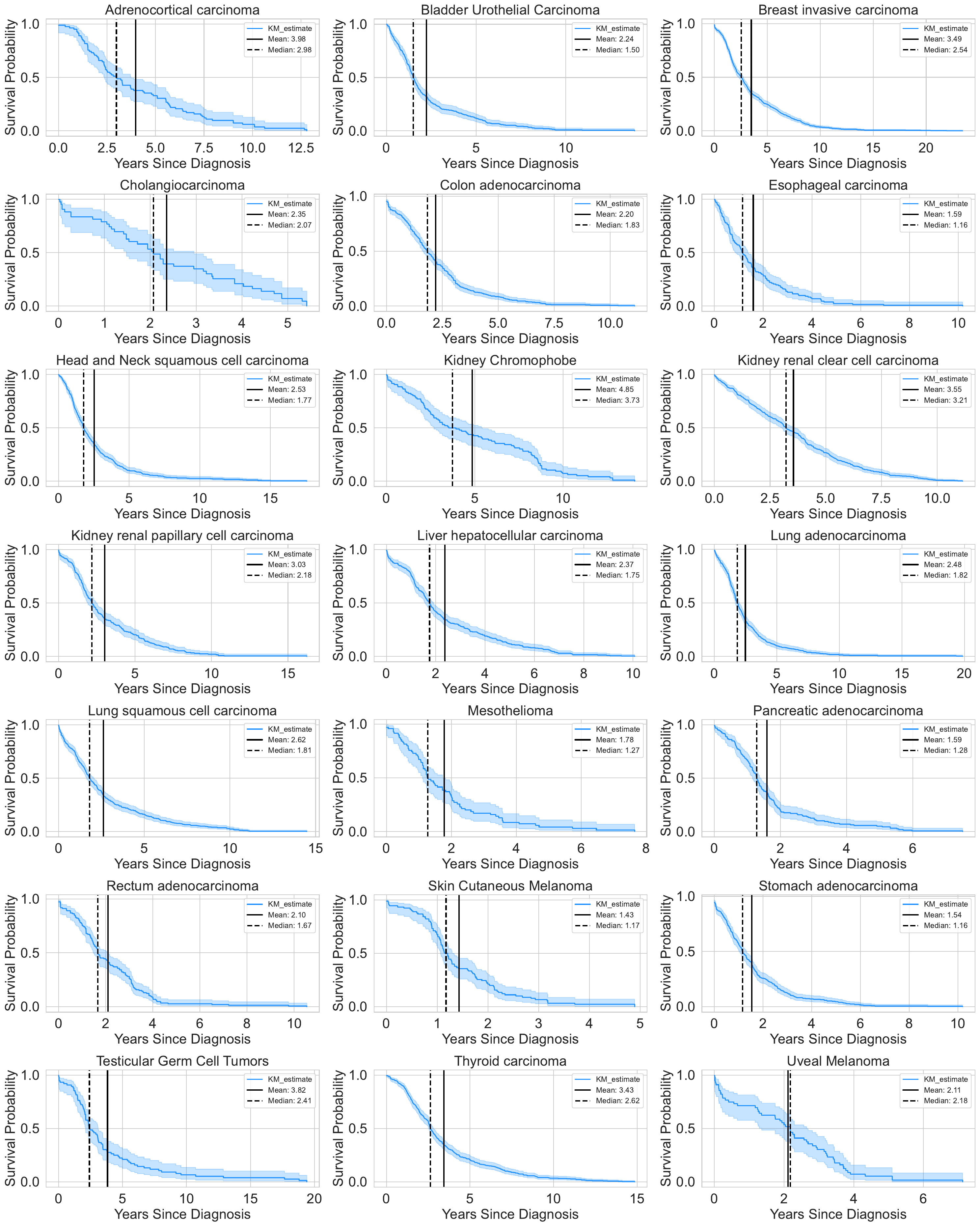}  
      \caption{
      Kaplan-Meier curves for all cancer types for prognosis assessment. 
      }
    \label{KM_plots}
\end{figure}

From Figure \ref{Fig2-Error-rates}(E) and the Kaplan-Meier curves (Figure \ref{KM_plots}), it is evident that the majority of cancer types included in our analysis have a mean survival time ranging between 1 and 4 years. Incorporating this information into the prompt enhanced the LLM's understanding of disease severity, thereby improving its ability to analyze and predict prognosis more effectively. As illustrated in Figure \ref{Fig4-Cot-Trace-Prog}, GPT-4o systematically evaluates all relevant factors for Stomach Adenocarcinoma by constructing a rubric to identify positive and negative indicators for prognosis. The model effectively analyzes the pathology report, considering key elements such as lymph node involvement. For instance, it correctly concludes that the absence of lymph node involvement is a favorable indicator for the patient. Additionally, it examines resection margins and determines that since the margins are intact post-surgery, this also suggests a positive prognosis. This structured reasoning approach demonstrates LLMs' capability to mimic clinical decision-making processes when provided with appropriate guidance.

%% file: 3_discussion.tex
\section{Discussion}

In this paper, we present a evaluation of the ability of recent large language models (LLMs) to assess cancer type, AJCC stage, and prognosis from pathology reports in a zero-shot setting. Our findings indicate that while LLMs excel in information extraction tasks, their performance declines in higher-level deduction-based reasoning tasks, such as determining cancer stage and prognosis. A potential enhancement to these tasks would be to reframe the first two problems—cancer type identification and AJCC staging—by incorporating multiple examples or shots within the prompt, similar to the approach used for the prognosis assessment task.

Additionally, we developed a robust evaluation framework that employs strict performance metrics and enforces a standardized output format, specifically JSON, facilitating eventual deployment in clinical settings. Furthermore, we present a methodology for instruction tuning LLMs, demonstrating that even smaller, open-source models can achieve performance on par with, or surpass, proprietary large LLMs across these three tasks.

Our analysis reveals that employing a self-generated chain-of-thought strategy significantly enhances model performance when used natively for inference. However, this approach can be computationally expensive during deployment due to the increased token usage in both inputs and outputs. Similarly, incorporating a few examples (or shots) in the system prompt improves predictions but suffers from the same limitation. Instruction tuning offers a practical solution by reducing inference time and computational costs, enabling hospitals to deploy specialized LLMs tailored for predefined tasks. These models are lightweight, require fewer resources, and are cost-effective to operate. Moreover, from a privacy standpoint, instruction-tuned models provide an additional advantage, as they can be deployed locally, ensuring that sensitive patient data is not shared with third-party services.

While our research demonstrates the potential of LLMs in medical applications, there remain significant hurdles to overcome. One common issue is that LLMs often confuse textually similar diseases, highlighting the need for more robust system prompts to mitigate such errors. A more pressing challenge lies in estimating the AJCC stage, where the best observed performance is around 87\%. Given the subjective nature of staging, establishing strict and standardized guidelines across diseases could facilitate large-scale data mining and improve model consistency in the future. In terms of prognosis assessment, while LLMs show promise, they are not yet ready for clinical deployment. Integrating additional data sources, such as surgical records, laboratory results, and other electronic medical record (EMR) data, could significantly enhance their performance. This multimodal approach could provide a more comprehensive understanding of patient conditions, bringing LLMs closer to practical applications in clinical settings.

A next step in research, incorporating clinical data directly into the prompts could help LLMs make more informed and accurate decisions. An ambitious but impactful extension of our work would involve extracting chain-of-thought reasoning traces from pathologists and using this data for large-scale instruction tuning. Such an approach could significantly enhance the models' ability to mimic expert reasoning, ultimately leading to higher performance metrics, particularly in challenging tasks like prognosis assessment.

As a future avenue of research, incorporating a retrieval-augmented generation (RAG) framework \cite{lewis2020retrieval} could significantly enhance the performance of open-source LLMs in these tasks. This approach would allow LLMs to access and reference hospital-specific guidelines, other pathology reports, and electronic medical record (EMR) texts, enabling them to make more informed decisions about staging and prognosis while maintaining privacy. Leveraging these diverse textual resources would provide the LLM with a broader and contextually richer knowledge base, improving its ability to generate accurate and reliable predictions. This strategy aligns with the goal of creating robust, specialized systems for clinical deployment.

In conclusion, LLMs demonstrate the potential to accurately identify cancer type, determine AJCC stage, and assess prognosis from pathology reports. In this study, we showcased various prompting methodologies and benchmarked both open-source and proprietary LLMs across a wide range of cancer types. Additionally, we developed two instruction-tuned models: one fine-tuned using OpenAI's platform and another based on a 8B open-source LLM. The latter offers a resource-efficient and privacy-preserving solution, making it well-suited for deployment in clinical environments with minimal computational overhead.

\begin{figure}[H]
      \centering
      \includegraphics[width=1\linewidth]{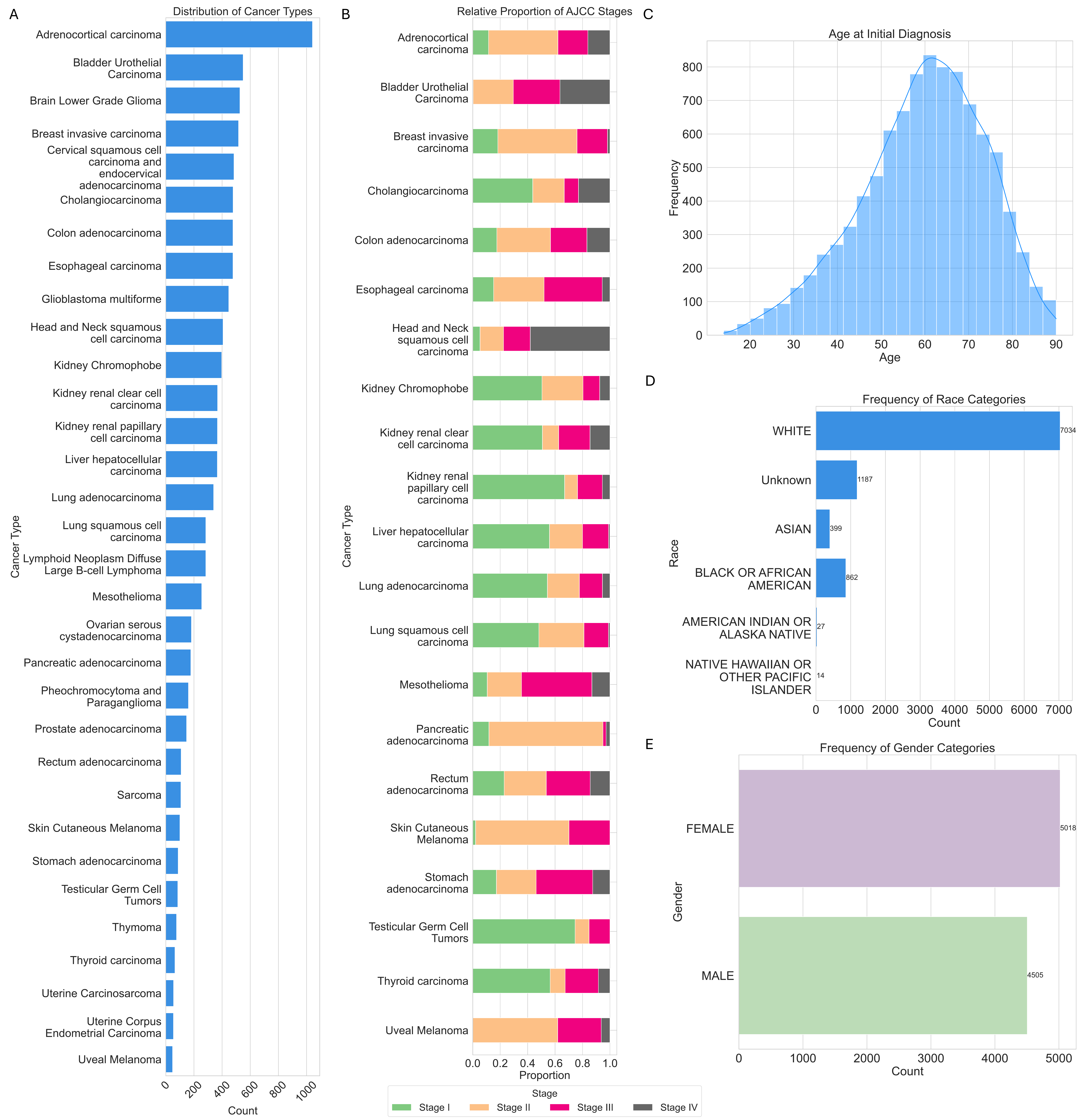} 
      \caption{Summary overview of dataset. (A) Number of data points for each cancer type. (B): Distribution of AJCC stages for each cancer type where staging is present. (C) Distribution of age of patients at the time of diagnosis. (D) Distribution of race of patients. (E) Distribution of gender of patients.}
    \label{Fig-dataset}
\end{figure}

%% file: 4_methods.tex
\section{Methods}
\label{Section-Methods}

\subsection{Data Source}

We created the dataset used in this study by combining two publicly available datasets, both derived from The Cancer Genome Atlas (TCGA) initiative \cite{weinsteinCancerGenomeAtlas2013}. 
The TCGA-Reports dataset comprises the text of more than 9,000 pathology reports, extracted from the original PDF reports and cleaned into a machine-readable dataset \cite{kefeliTCGAReportsMachinereadablePathology2024}. 
The TCGA Pan-Cancer Clinical Data Resource (TCGA-CDR) contains clinical data including survival data for multiple outcomes measures \cite{liuIntegratedTCGAPanCancer2018}. These two datasets were merged using the unique sample-level TCGA barcodes. The final dataset comprised 9523 unique samples.

From Figure \ref{Fig-dataset}(A), we observe the distribution of cancer types across the dataset, which includes 32 distinct types of cancers. For the cancer staging task, we considered only the high-level stage groups (e.g. Stage I, Stage II, etc.) and grouped sub-stages together. For example, Stage IIa and Stage IIb were both considered Stage II.
Not all cancer types have stage data available, and relative frequencies of each stage vary between cancer types as shown in Figure \ref{Fig-dataset}(B).

For the prognosis task, disease-specific survival (DSS) was selected as the primary endpoint, with its distribution shown in Figure \ref{Fig2-Error-rates}(E). DSS was chosen because pathology reports focus exclusively on the specimens examined, making DSS a more specific and relevant measure of prognosis for this disease process compared to broader metrics like overall survival. Prognostic categories were binarized based on the mean DSS within each cancer type to account for the significant variability in survival across different cancer types. This approach ensures that the classification aligns with the unique characteristics of each cancer type.

From Figure \ref{Fig-dataset}(C), we observe the distribution of age at initial diagnosis for all data points, with the mean age falling between 60 and 70 years. The dataset is skewed with respect to race but does include representation from minority groups, as shown in Figure \ref{Fig-dataset}(D). Additionally, the dataset is relatively balanced in terms of gender distribution, with a slightly higher number of female patients compared to male patients. The dataset was randomly split into training, validation, and testing sets according to an 80:10:10 ratio, stratified on cancer type to ensure similar class balance. The complete data records are made public.

\subsection{Methodology for Cancer Type Identification}

All the LLMs, including the instruction-tuned models, were subjected to the same standardized prompt. Prompt engineering and selection, as noted in \cite{sahoo2024systematic}, can be challenging and time-consuming, so we opted for a minimal and efficient approach. Each LLM was provided with both a system and a user prompt.

For the system prompt, we instructed the model to act as a highly knowledgeable pathology AI assistant, which helps align its internal representations and enhances its ability to generate accurate responses. Additionally, we constrained the model to select answers only from the cancer types present in the dataset. This restriction prevents the LLM from generating irrelevant cancer types and proves beneficial for consistent and effective data mining.

We also instructed the LLMs to avoid printing any extraneous text, addressing their inherent verbosity. Additionally, we explicitly directed the models not to fabricate information, thereby minimizing the likelihood of hallucination. Finally, we required the models to generate only the JSON object as the output \footnote{As of February 26, 2025, OpenAI provides structured output capabilities on its API platform. However, this feature is not yet available for open-source models.}, without providing any explanations or additional context, ensuring concise and standardized responses. An example of the prompt can be seen in Figure \ref{Fig7-Prompt-Cancer Type}, in the Appendix.

\subsection{Methodology for AJCC Stage Identification}

Identifying the AJCC stage is undoubtedly a more challenging task that requires a level of reasoning. Initially, attempts to have the LLMs predict the AJCC stage in a purely zero-shot setting were unsuccessful. Various prompting techniques were explored, including asking the models to first predict the TNM staging and then derive the AJCC stage, but these approaches also failed to yield accurate results. Ultimately, employing a self-generated chain-of-thought \cite{wei2022chain} technique proved effective, enabling the LLMs to reason through the task and arrive at more accurate answers.

From Figure \ref{Fig4-Framework}(B), we observe that, similar to the cancer type identification task, a zero-shot prompting framework is employed, but with the addition of a self-generated chain-of-thought reasoning process that incorporates four AJCC stage options. As shown in Figure \ref{Fig8-Prompt-AJCC Stage} in the Appendix, we provide a well-structured template to guide the LLM in explaining its reasoning before presenting the final answer in JSON object format. Furthermore, the user prompt is restructured as a multiple-choice question, which has been shown to enhance performance by providing a clear decision framework for the model.

During inference, the system and user prompts for the instruction-tuned LLM do not include any chain-of-thought reasoning. Instead, the model is designed to operate in a purely zero-shot framework, directly generating outputs without any intermediate reasoning steps.

\subsection{Methodology for Prognosis Assessment}

\begin{figure}[H]
      \centering
      \includegraphics[width=1\linewidth]{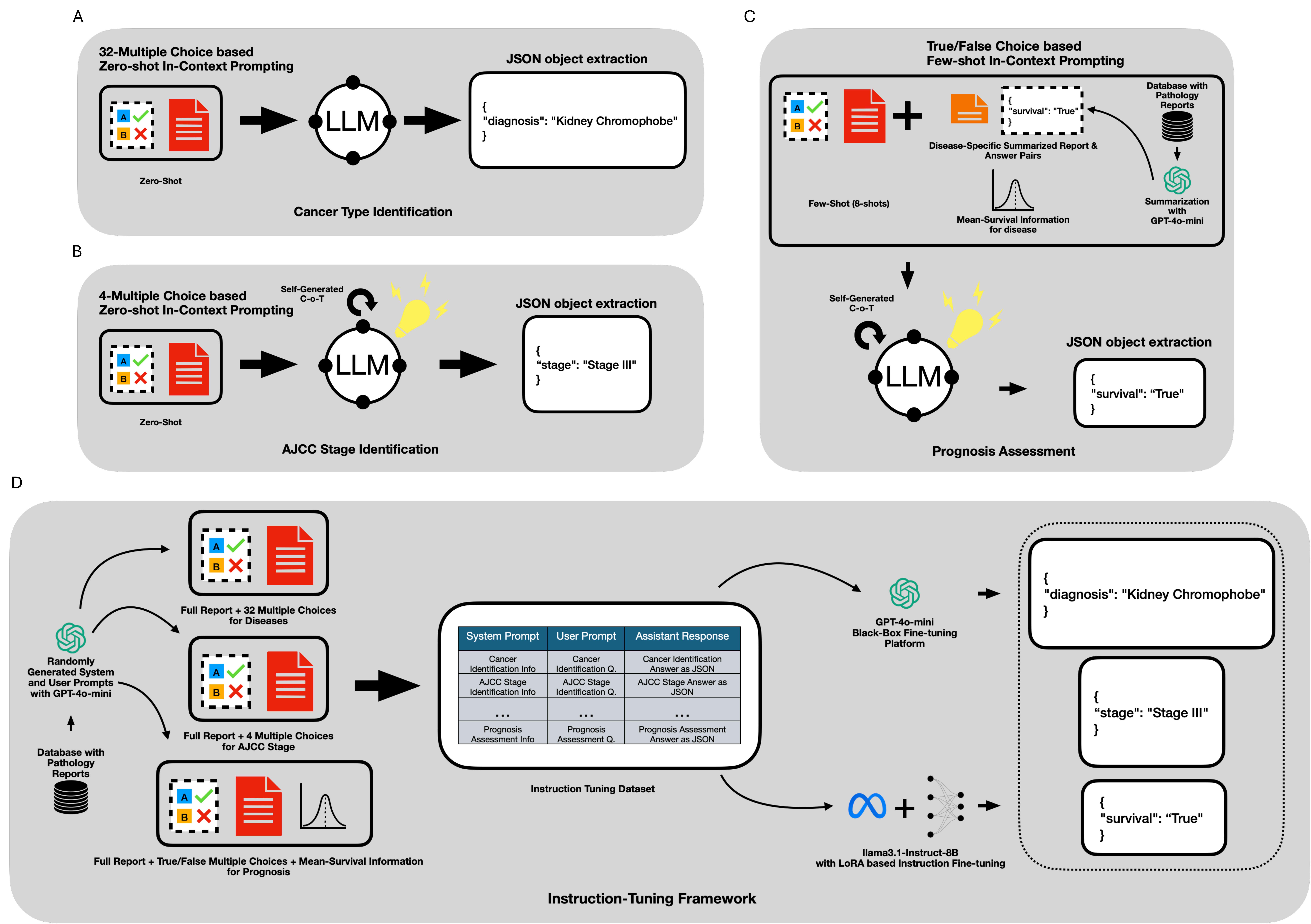}  
      \caption{Summary overview of methodology. (A) Cancer Type Identification Methodology (B): AJJC Cancer Identification Methodology. (C) Prognosis Assessment Methodology (D) Instruction-tuning Methodology}
    \label{Fig4-Framework}
\end{figure}

For the Prognosis Assessment task, similar to the stage identification task, a zero-shot prompting strategy alone proved ineffective, and even a zero-shot self-generated chain-of-thought approach failed to produce reliable results. To address this challenge, we restructured the problem by identifying a key limitation: the LLM lacked a reference point for distinguishing between good and poor prognoses for a given cancer type. While it may seem counterintuitive that large-scale LLMs lack this world knowledge, empirical observations suggest that they tend to predict extreme prognoses—either overly optimistic or highly severe—rather than producing a balanced assessment. To overcome this (as shown in Figure \ref{Fig4-Framework}(C)), we first extracted the mean disease-specific survival (DSS) time for each cancer type and included it in the prompt. Since DSS values provide a contextual anchor for prognosis estimation, similar summary statistics could also be sourced from the literature, ensuring that this approach remains applicable even when working with new datasets where the mean DSS is unknown. Additionally, we provided the LLM with four examples of good prognoses and four examples of poor prognoses, enabling it to better contextualize and evaluate prognosis predictions. In this context, a good prognosis is defined as a patient surviving beyond the mean disease-specific survival (DSS) time, while a bad prognosis is when survival falls below the mean DSS time. To further enhance the model's performance, we also incorporated a self-generated chain-of-thought approach in the prompt. This strategy encourages the LLM to reason through the provided information systematically, improving its ability to make accurate prognosis predictions.

The few-shot framework presented challenges due to the length of the pathology reports, as including 8 full reports as examples often exceeded token limits and context length constraints. To address this issue, we summarized the reports using the GPT-4o-mini model, reducing their length while retaining essential information. These summaries were then included in the prompt, enabling the framework to remain within token limits while still providing relevant examples to guide the model's predictions effectively.  The example prompt can be seen in Figure \ref{Fig9-Prompt-Prognosis}, in the Appendix. 

During the evaluation or inference phase, the 8 examples (4 positive and 4 negative) are randomly selected from the training data and included in the prompt for guiding the model. However, when evaluating the instruction-tuned LLM, no chain-of-thought reasoning or few-shot examples are included. Instead, the instruction-tuned model operates in a pure zero-shot framework, with only the mean DSS time provided as context to support its predictions.

\subsection{Instruction-Tuning Framework}

To build instruction-tuned models, we begin by synthetically generating question-and-answer pairs along with their corresponding system prompts. This is achieved by leveraging the training and validation datasets and using GPT-4o-mini to create textually diverse but semantically equivalent system and user prompts. These prompts are paired with the pathology reports and their respective answers. This approach produces a highly heterogeneous set of questions, allowing the LLM to be exposed to a broader range of variations during training. As a result, the model becomes more robust and better equipped to handle diverse queries in a clinical context.

To clarify, the content of the pathology reports remains unchanged; only the questions and system prompts are modified to avoid inducing incorrect hallucinations. These synthetically generated question-and-answer pairs are created for all three tasks using the training and validation datasets. This process results in a total of 17,344 question-and-answer pairs for training and 2,151 pairs for validation, providing a diverse and comprehensive dataset to effectively train the instruction-tuned models. The framework can be seen in Figure \ref{Fig4-Framework}(D). 

For the \textbf{Path-GPT-4o-mini-FT}, we directly utilize this dataset and fine-tune the model using the OpenAI Fine-Tuning platform. While the internal training strategy of the platform is not disclosed, we configured the training process with a learning rate multiplier of 0.3 and trained the model for approximately 17,000 steps. The number of steps was determined arbitrarily, as the internal mechanisms of OpenAI's fine-tuning platform remain unknown. The convergence of the training loss curve is shown in Figure \ref{Fig2-Error-rates}(D).

The training process for \textbf{Path-llama3.1-8B} involves specific considerations. We use the Llama3.1-8B-Instruct model as the base model for instruction tuning. Llama3.1 was chosen over Llama3 because, empirically, it demonstrates superior instruction-following capabilities. However, during direct inference and in-context prompting, Llama3 was observed to perform better on the tasks in this project, indicating that the choice of model depends on the specific use case and whether instruction tuning is applied.

To perform instruction tuning, we use Low-Rank Adaptation (LoRA) \cite{hu2022lora} and fine-tune specific components of the model. Specifically, we adjust the Query Projection module to map the input embeddings into the query space and the Key Projection module to map the input embeddings into the key space. These adjustments are crucial for computing the attention weights within the LLM effectively. Additionally, we fine-tune the Value Projection module, which is responsible for extracting relevant information from the pathology reports.

Alongside these, we also fine-tune certain intermediate projectors to better align the model's internal representations with the specific tasks and tune the Output Projection module to enhance task-specific outputs. This targeted approach allows us to adapt the model efficiently while preserving its overall capacity and ensuring it performs optimally for the given problem.

We use a LoRA rank of 16 and a LoRA alpha of 16 to implement rank-stabilized LoRA. These higher values were chosen because the task is particularly challenging, and this configuration was empirically determined to be optimal, balancing performance without leading to overfitting. To expedite training, we set the dropout rate to 0 and employed gradient checkpointing to reduce memory usage, enabling efficient fine-tuning on resource-constrained hardware.

The instruction tuning was conducted for 6,000 steps, as shown in Figure \ref{Fig2-Error-rates}(E), using a 4-bit quantized version of the Llama3.1-Instruct model. The maximum sequence length during training was set to 4,096 tokens to accommodate longer input contexts. The model was fine-tuned with a learning rate of $3e^{-4}$ using the \textit{adamw\_8bit} optimizer, which balances performance and memory efficiency, making it well-suited for resource-constrained training setups. The rest of training parameters are available in the code. 